\documentclass{article}
\pdfoutput=1
\usepackage{arxiv}
\setcitestyle{authoryear,open={(},close={)}}
\renewcommand{\cite}{\citep}

\usepackage[utf8]{inputenc} %
\usepackage[T1]{fontenc}    %
\usepackage{hyperref}
\usepackage{url}
\usepackage{booktabs}       %
\usepackage{amsfonts}       %
\usepackage{nicefrac}       %
\usepackage{microtype}      %
\usepackage{dsfont}
\usepackage{graphicx}
\usepackage{amsmath}
\usepackage{mathtools}
\usepackage{amssymb}
\usepackage{amsthm}
\usepackage{cases}
\usepackage{color}
\usepackage{textcomp}
\usepackage{xcolor}
\usepackage{caption}
\usepackage{bbm}
\usepackage{comment}
\usepackage{xspace}
\usepackage{multirow}
\usepackage{subcaption}
\usepackage{authblk}
\usepackage{xspace}
\usepackage{algpseudocode}
\usepackage{adjustbox}
\usepackage{wrapfig}

\usepackage{microtype}
\usepackage{graphicx}
\usepackage{booktabs} %

\usepackage{hyperref}

\usepackage{amsmath}
\usepackage{amssymb}
\usepackage{mathtools}
\usepackage{amsthm}
\usepackage{mleftright}

\usepackage[capitalize,noabbrev]{cleveref}

\theoremstyle{plain}

\theoremstyle{remark}

\usepackage{amsfonts,bm}

\def\eqref#1{equation~\ref{#1}}

\def\1{\bm{1}}

\DeclareMathAlphabet{\mathsfit}{\encodingdefault}{\sfdefault}{m}{sl}
\SetMathAlphabet{\mathsfit}{bold}{\encodingdefault}{\sfdefault}{bx}{n}

\DeclarePairedDelimiterX{\KLdivx}[2]{\big(}{\big)}{%
  #1\;\delimsize\|\;#2%
}

\usepackage{amsthm}

\newtheorem{property-non}{Property}

\hypersetup{
    colorlinks,
    citecolor=[rgb]{0.00, 0.45, 0.70}, %
    linkcolor=[rgb]{0.01, 0.62, 0.45}, %
    urlcolor=[rgb]{0.68, 0.45, 0.63} %
}
\definecolor{green1}{rgb}{0.01, 0.62, 0.45}
\definecolor{blue1}{rgb}{0.00, 0.45, 0.70}
\definecolor{blue2}{rgb}{0.612, 0.8, 0.902}
\definecolor{purple1}{rgb}{0.68, 0.45, 0.63}
\definecolor{red1}{rgb}{1.0, 0.1, 0.3}
\definecolor{red2}{rgb}{1.0, 0.77, 0.80}

\usepackage{todonotes}
\usepackage{enumitem}

\usepackage{tikz}
\usepackage{xcolor}
\usepackage{graphicx}
\usetikzlibrary{arrows.meta,decorations.pathreplacing,calc,shapes.geometric,positioning,shapes.misc}
\usepackage{amsmath}
\usepackage{amsthm}
\usepackage{amssymb}
\usepackage[bbgreekl]{mathbbol}

\DeclareMathSymbol\drule  \mathord{bbold}{"01}

\usepackage{bbm}
\usepackage[ruled]{algorithm2e}

\usepackage{wrapfig}
\usepackage{graphicx}
\usepackage{caption}
\usepackage{titletoc}
\usepackage{placeins}
\usepackage{graphicx}
\usepackage{placeins}
\usepackage[most]{tcolorbox} %
\usepackage{pdflscape}

\usepackage{tabularx}
\usepackage{siunitx}
\usepackage{soul}

\renewrobustcmd{\bfseries}{\fontseries{b}\selectfont}
\renewrobustcmd{\boldmath}{}
\newrobustcmd{\B}{\bfseries}

\newtcolorbox{qualitativeBox}{
  colback=gray!10, %
  colframe=gray!20, %
  rounded corners, %
  boxrule=0.5pt, %
  left=10pt, %
  right=10pt, %
  top=5pt, %
  bottom=5pt, %
  boxsep=5pt, %
}
\newtcolorbox{ourMethodBox}{
  colback=blue1!10, %
  colframe=gray!20, %
  rounded corners, %
  boxrule=0.5pt, %
  left=10pt, %
  right=10pt, %
  top=5pt, %
  bottom=5pt, %
  boxsep=5pt, %
}
\newtcolorbox{theirMethodBox}{
  colback=red1!10, %
  colframe=gray!20, %
  rounded corners, %
  boxrule=0.5pt, %
  left=10pt, %
  right=10pt, %
  top=5pt, %
  bottom=5pt, %
  boxsep=5pt, %
}

\usepackage[capitalize,noabbrev]{cleveref}
\usepackage{xspace}
\usepackage[table]{xcolor}
\NewDocumentCommand{\update}{ mO{} }{\textcolor{blue}{{{#1}}}}
\newcommand{\toolName}{SSA\xspace}

\NewDocumentCommand{\yuzhen}{ mO{} }{\textcolor{red}{\textsuperscript{\textit{yuzhen}}\textsf{\textbf{\small[#1]}}}}
\definecolor{posgreen}{RGB}{34,139,34}
\definecolor{colorFULL}{HTML}{FFF4E5} 
\definecolor{colorX8}{HTML}{EAF4FF}  
\definecolor{colorX16}{HTML}{E8F5E9} 
\definecolor{colorX32}{HTML}{F5EFFF} 

\usepackage{lineno}

\definecolor{darkblue}{rgb}{0, 0, 0.5}
\hypersetup{colorlinks=true, citecolor=darkblue, linkcolor=darkblue, urlcolor=darkblue}

\begin{document}

\title{Simplified Sparse Attention via Gist Tokens}
\author{
    \large{Yuzhen Mao, ~Michael Y. Li, ~Emily B. Fox}\linebreak
    \large{Stanford University}\linebreak
        {\texttt{\{yuzhenm,michaelyli,ebfox\}@stanford.edu}}
}

\date{}

\maketitle

\begin{abstract}
{
Sparse attention can reduce the cost of long-context inference, but most variants introduce new architectural components. 
We introduce Simplified Sparse Attention (\toolName{}), a simpler approach to sparse attention that requires no architectural changes.
Concretely, we first perform continued pretraining  on sequences interleaved with gist tokens. 
We optimize the standard next-token loss as usual, but the gist tokens use an attention mask to restrict what parts of the context the language model can attend to; this teaches the model to pack each chunk's important information into the gist tokens. 
At inference time, \toolName{} scores chunks via attention between the current query and the small set of gist tokens, \emph{selectively unfolding} the top-$k$ chunks by reintroducing their corresponding raw tokens. 
Since the query is scored only against the gist tokens, we avoid the memory-bandwidth cost associated with naive scoring against the full KV cache, without requiring the auxiliary KV cache approach used by sparse attention methods. 
On LongBench, \toolName{} consistently outperforms compression and inference-time sparse-attention baselines under the same compression ratio. More strikingly, in retrieval-augmented generation, \toolName{} can even outperform the continued-pretrained full-attention model by over 5.7 points. 
We attribute this to the ability of \toolName{}'s selective unfolding, which concentrates attention on the query-relevant chunks and effectively filters out noise. Beyond accuracy, \toolName{} also improves efficiency: its decoding latency
remains nearly flat as context length increases, yielding up to a
3.37$\times$ end-to-end decoding speedup over Flash-Decoding,
while matching and eventually outperforming the FlashAttention baseline during prefill.
\toolName{} further extends to a hierarchical gist-of-gist variant (H-\toolName{}) that achieves log-linear decoding complexity while maintaining or improving accuracy at high compression ratios up to 32$\times$. The code is available at~\url{https://github.com/yuzhenmao/simplified-sparse-attention/}.
}
\end{abstract}

\section{Introduction}
\label{sec:intro}
Long-context modeling has emerged as a critical capability for next-generation large language models (LLMs), driven by diverse real-world applications including in-depth reasoning~\citep{zelikman2022star, guo2025deepseek}, repository-level software engineering~\citep{zhang2023repocoder,yang2024sweagent}, and multi-turn autonomous agent systems~\citep{alphaevolve2025,liu2026skydiscover}. However, as context lengths grows, the attention computation becomes increasingly dominant in the overall cost, presenting significant obstacles for both training and inference.


\input{sections/0_figure.tex}

Sparse attention offers a promising way to mitigate this bottleneck at inference time. 
By restricting each query to attend to only a subset of keys, sparse methods can substantially reduce the cost of attention while preserving model capabilities. Existing approaches can be broadly categorized along two axes: methods that apply sparsity at inference time and methods that also introduce sparsity during training.

On the inference side, methods such as H$_2$O~\citep{zhang2023h2o}, StreamingLLM~\citep{xiao2023efficient}, and Quest~\citep{tang2024quest} apply sparsity during decoding by evicting or selecting KV-cache entries. While effective at reducing memory footprint, these approaches are applied post hoc to full-attention pretrained models, treating sparsity as an approximation rather than allowing the model to learn input-dependent sparse patterns during pretraining. 
On the training side, recent methods including Native Sparse Attention~\citep[NSA;][]{yuan2025native}, Deepseek Sparse Attention~\citep[DSA;][]{liu2025deepseek} and MoBA~\citep{lu2025moba} have demonstrated that incorporating sparsity during pretraining can preserve or even improve model quality while achieving substantial speedups.
This type of idea has also been explored in post-training in Neural Garbage Collection~\citep{li2026neuralgarbagecollectionlearning} which trains a language model to evict from its own KV cache using reinforcement learning.
However, existing training-based approaches often require additional architectural components or specialized mechanisms. This raises a natural question: can we obtain the benefits of training-based sparse attention without modifying the model architecture?

To design an architecture-free sparse attention method, we take inspiration from gist tokens \citep{mu2023learning, chevalier2023adapting, yang2025kvlink, zhang2024long, petrov2025long, deng2025unigist}. 
The idea is to insert a small number of special tokens---gist tokens---at fixed positions in the sequence, partitioning the context into chunks.
Training uses the standard next-token prediction objective but with a change to the attention mask:
tokens in a chunk can attend to earlier raw tokens only through the preceding chunk's gist tokens, not directly. 
Therefore, to minimize next-token loss under this constraint, the model must learn to pack the information needed for future predictions into the gists. 
Compression therefore emerges as a consequence of end-to-end optimization under constraints~\citep{li2026neuralgarbagecollectionlearning}, with no auxiliary loss, no new parameters, and no architectural additions.

To that end, we introduce \textbf{Simplified Sparse Attention (\toolName)}, which interleaves gist tokens during continued pretraining and then uses these gist tokens to perform sparse attention during decoding. 
Concretely, after compressing the context into interleaved gist tokens, we perform top-$k$ selection over these gist tokens based on their attention scores with respect to the current query, and then \emph{selectively unfold} the corresponding chunks by reintroducing their raw tokens into the context. 

Remarkably, we find that an entirely pretraining-based intervention suffices to make this top-$k$ selection procedure highly effective.
In particular, under a fixed token budget, this selective unfolding strategy significantly improves accuracy compared to standard compression baselines, including both append-only and interleaved gist methods, as well as inference-only sparse attention techniques.
To explicitly teach the model to adapt to sparse attention, we can also {incorporate the selective unfolding mechanism into the training process}.
Importantly, \toolName{} is lightweight and simple to implement: it re-purposes a standard training procedure, continued pre-training, into a general purpose recipe for turning any full-attention architecture into a sparse attention method.

We additionally observe that gist tokens can also compress \emph{other} gist tokens; in principle this procedure can continue \emph{ad infinitum}.
By hierarchically compressing gist tokens into higher-level summaries (meta-gist), we obtain a multi-resolution representation of the context.
The selective unfolding mechanism can then be applied in a coarse-to-fine manner: starting from the highest level of abstraction, the model progressively identifies the most relevant sub-contexts, unfolding from coarse summaries down to the original raw tokens. 
This hierarchical design captures information at multiple levels of granularity while maintaining computational efficiency that scales log-linearly with context length.

\section{Problem Setup}
\label{sec:problem}

We consider autoregressive language modeling, where a model predicts the next token $x_t$ given all preceding tokens $\mathbf{x}_{<t} = (x_1, \dots, x_{t-1})$. The prediction is computed as:
\begin{equation}
    P(x_t \mid \mathbf{x}_{<t}) = \operatorname{Softmax}\bigl(\mathbf{W}_{\text{vocab}} \cdot \mathbf{h}_t\bigr),
\end{equation}
where $\mathbf{W}_{\text{vocab}} \in \mathbb{R}^{V \times d}$ is the output projection matrix and $\mathbf{h}_t$ is the hidden state at position $t$. In the standard Transformer, $\mathbf{h}_t$ is computed via full attention over all preceding key-value pairs $(\mathbf{K}_{<t}, \mathbf{V}_{<t})$:
\begin{equation}\label{eq:atten}
    \mathbf{h}_t = \operatorname{Attention}(\mathbf{q}_t, \mathbf{K}_{<t}, \mathbf{V}_{<t}) = \operatorname{softmax}\!\left(\frac{\mathbf{q}_t \mathbf{K}_{<t}^\top}{\sqrt{d}}\right) \mathbf{V}_{<t}.
\end{equation}
As the context length $t$ grows, the cost of computing $\mathbf{h}_t$ and the memory required to store $(\mathbf{K}_{<t}, \mathbf{V}_{<t})$ both scale linearly per step and quadratically over the full sequence, making long-context modeling computationally prohibitive.

\paragraph{Goal.}
Our goal is to construct, at each decoding step $t$, a compact context $\mathcal{V}_t$ with $|\mathcal{V}_t| \ll t$ key-value pairs, such that the model's prediction quality is preserved:
\begin{equation}
    \mathbf{h}_t \approx \operatorname{Attention}(\mathbf{q}_t,\; \mathbf{K}_{\mathcal{V}_t},\; \mathbf{V}_{\mathcal{V}_t}).
\end{equation}
We seek a method that satisfies three desiderata. First, the compact context should be \emph{query-adaptive}: different queries may require different subsets of the history, and the selection should be dynamic rather than fixed. Second, the selection mechanism should be \emph{end-to-end trainable}, so that the model can learn to identify relevant context during pretraining or finetuning, rather than relying on heuristic or non-differentiable operations applied post hoc. Third, the approach should require \emph{no architectural modifications} to the standard Transformer, enabling straightforward integration with existing pretrained models and infrastructures.


\section{Method}
\label{sec:method}
{We present Simplified Sparse Attention (\toolName{}), a framework that bridges learnable gist-based context compression with sparse attention through a selective unfolding mechanism. The key idea is that interleaved gist tokens, trained to summarize local chunks, can simultaneously serve as \emph{learned routing signals} for identifying the $k$ most relevant sub-contexts. When a chunk is deemed relevant, its raw tokens are selectively reintroduced into the attention context, thereby recovering fine-grained detail precisely where needed. This results in a bounded context size of $|\mathcal{V}_t| = k \cdot (1 + L)$, where $L$ is the chunk length and is independent of the full sequence length. The framework further supports two complementary stages: continued pretraining alone already enables effective selective unfolding at inference time, while optional selective finetuning further improves performance, making \toolName{} both effective and flexible in practice.}

{Our framework is inspired by how humans process long contexts. When reading a novel or navigating a large codebase, readers do not retain every detail uniformly. Instead, they compress earlier content into high-level summaries---the gist---which is usually sufficient for ongoing comprehension. Only when a specific detail becomes necessary (e.g., a character’s exact words or a function’s signature) do they selectively revisit the relevant passage. \toolName{} instantiates this strategy: gist tokens maintain compact, continuously updated summaries of prior context, while selective unfolding enables targeted recovery of fine-grained information precisely when needed.}

We first review the sparse attention formulation and the interleaved gist token setup (\S\ref{sec:prelim}). We then describe the core selective unfolding mechanism at a single level of compression (\S\ref{sec:selection}--\S\ref{sec:hybrid}), followed by the training procedure (\S\ref{sec:training}). Next, we show how the framework naturally extends to a hierarchical gist-of-gist structure for log-linear complexity (\S\ref{sec:hierarchical}). Finally, we describe the corresponding kernel design (\S\ref{sec:kernel}).

\subsection{Preliminaries}
\label{sec:prelim}
\paragraph{Sparse Attention.}
In standard attention (\eqref{eq:atten}), each query $\mathbf{q}_t \in \mathbb{R}^{d}$ attends to all preceding key-value pairs. As the sequence length grows, this scales quadratically in $t$. Sparse attention methods reduce this cost by restricting each query to attend to a subset $\mathcal{S}_t \subset \{1, \dots, t\}$ of positions:
\begin{equation}
    \mathbf{o}_t = \sum_{i \in \mathcal{S}_t} \alpha_{t,i} \, \mathbf{v}_i, \quad \alpha_{t,i} = \frac{\exp(\mathbf{q}_t^\top \mathbf{k}_i / \sqrt{d})}{\sum_{j \in \mathcal{S}_t} \exp(\mathbf{q}_t^\top \mathbf{k}_j / \sqrt{d})},
\end{equation}
where $|\mathcal{S}_t| \ll t$. The core challenge lies in how to construct $\mathcal{S}_t$: fixed patterns such as sliding windows or stride-based selection impose structural priors that may miss relevant tokens, while dynamic methods that adapt $\mathcal{S}_t$ to each query require an efficient mechanism for estimating token relevance without computing full attention.
\paragraph{Interleaved Gist Tokens.}
Following prior work on gist-based context compression~\citep{mu2023learning, zhang2024long, petrov2025long, deng2025unigist}, we partition the input sequence $\mathbf{X} = (x_1, \dots, x_n)$ into $M = \lceil n/L \rceil$ non-overlapping chunks of length $L$: $\mathcal{C} = \{C_1, \dots, C_M\}$, where $C_m = (x_{(m-1)L+1}, \dots, x_{mL})$. A learnable gist token $g_m$ is appended to each chunk, yielding the augmented sequence:
\begin{equation}
    \tilde{\mathbf{X}} = (C_1, \; g_1, C_2, \; \dots, \; C_M, g_M).
\end{equation}
{As illustrated in Figure~\ref{fig:overview}, a causal mask $\mathbf{M}_{\text{gist}}$ is applied such that tokens within chunk $C_m$ can attend to tokens within $C_m$ and to all preceding gist tokens $g_{<m}$, while tokens following $g_m$ cannot attend to raw tokens within $C_m$'s scope---only to $g_m$ and the preceding gist tokens $g_{<m}$. This creates an information bottleneck that forces each gist token to learn a compressed representation of its corresponding chunk. After a forward pass, the embedding $\mathbf{g}_m$ summarizes $C_m$, and its associated KV pair $(\mathbf{k}_{g_m}, \mathbf{v}_{g_m})$ can then replace the full KV-cache of $C_m$ during decoding.}

\subsection{Gist-Based Relevance Scoring}
\label{sec:selection}
The central observation of this work is that interleaved gist tokens, beyond their role as compressed summaries, can naturally serve as \emph{selection proxies} for their associated chunks. Because each gist token $g_m$ encodes the semantic content of chunk $C_m$, the attention affinity between a query and a gist token provides a direct, learned estimate of the chunk's relevance.

Concretely, at each decoding step $t$, given the current query $\mathbf{q}_t$, we compute a relevance score for each chunk $C_m$ via the dot product between $\mathbf{q}_t$ and the gist key $\mathbf{k}_{g_m}$:
\begin{equation}
\label{eq:score}
    s_{t,m} = \mathbf{q}_t^\top \mathbf{k}_{g_m}.
\end{equation}
{This scoring mechanism is conceptually related to the gating function in MoBA~\citep{lu2025moba}, which computes the affinity between a query and the \emph{mean-pooled} keys of each block: $s_i = \mathbf{q}^\top \overline{\mathbf{K}}_{[I_i]}$. However, there is a crucial difference. Mean pooling is a fixed, non-parametric operation: the block representation is simply the arithmetic average of the raw keys, which may wash out semantically important signals and has no learnable parameters. In contrast, gist tokens are \emph{learned} representations optimized through training. This enables two advantages: (i)~the gist token can capture rich semantic content beyond what a simple average reflects, and (ii)~gradients flow through the gist tokens, allowing the model to jointly optimize the \emph{quality of compression} and the \emph{accuracy of selection}.}
 
This scoring is also related to NSA's compressed attention branch~\citep{yuan2025native}, which generates coarse-grained summary tokens via a dedicated MLP and uses the resulting attention scores to identify important fine-grained chunks. Our approach achieves a similar effect without requiring additional architectural components: the gist tokens are a part of the standard input sequence and reuse the existing attention mechanism.
\subsection{Selective Unfolding and Hybrid Attention}
\label{sec:hybrid}
Given the relevance scores, we select the top-$k$ chunks with the highest scores:
\begin{equation}
\label{eq:topk}
    \mathcal{I}_t = \operatorname{Top\text{-}k}\bigl(\{s_{t,m}\}_{m=1}^{M}\bigr).
\end{equation}
For selected chunks $m \in \mathcal{I}_t$, we \emph{unfold} the compressed representation by reintroducing both the gist KV pair  $(\mathbf{k}_{g_m}, \mathbf{v}_{g_m})$ and the full raw KV pairs $(\mathbf{K}_{C_m}, \mathbf{V}_{C_m})$ into the attention context. Unselected chunks are \emph{entirely excluded} from the decoding context. The hybrid key and value matrices are thus constructed as:
\begin{equation}
\label{eq:hybrid_kv}
    \mathbf{K}_{\text{hybrid}} = \bigcup_{m \in \mathcal{I}_t} \bigl(\mathbf{k}_{g_m} \cup \mathbf{K}_{C_m}\bigr), \quad
    \mathbf{V}_{\text{hybrid}} = \bigcup_{m \in \mathcal{I}_t} \bigl(\mathbf{v}_{g_m} \cup \mathbf{V}_{C_m}\bigr).
\end{equation}
The output is then computed via standard attention:
\begin{equation}
\label{eq:output}
    \mathbf{h}_t = \operatorname{Attention}(\mathbf{q}_t,\; \mathbf{K}_{\text{hybrid}},\; \mathbf{V}_{\text{hybrid}}).
\end{equation}
It is worth noting that we skip selective unfolding in the first attention layer, as all gist tokens share identical input hidden embeddings at this stage, rendering top-$k$ selection uninformative.

We experimented with several variants of the hybrid context, including (i)~retaining gist tokens for all chunks (both selected and unselected) as compact surrogates, and (ii)~using only the raw tokens from selected chunks without any gist tokens in the context. Empirically, we found that attending exclusively to the selected gist-chunk pairs yields the best performance (see \S\ref{sec:ablation}). We attribute this to the fact that, under a fixed token budget, replacing unselected gist tokens with unfolded raw tokens from selected gists provides a more informative attention context. The gist tokens of selected chunks are still retained alongside their raw tokens, as they provide complementary compressed information that captures cross-token patterns within the chunk.

During the experiments, rather than using a fixed Top-$k$ across all settings, we set the $k$ adaptively based on the sequence length, compression ratio, and GQA group structure, ensuring a consistent compression ratio across different configurations. Details of the adaptive top-$k$ formula are provided in Appendix~\S\ref{exp:topk}.

A practical benefit of routing context through gist tokens is that decoding only needs the gist KV cache to be resident, rather than the full per-token KV cache of the preceding context. This differs from many sparse attention methods: although they attend to only a selected subset of tokens, they often still need access to the full KV cache to score or identify which tokens should be selected. As a result, these methods either load the full cache during decoding or introduce an auxiliary module for token scoring. DSA~\cite{liu2025deepseek}, for example, creates a separate auxiliary indexer for this purpose. By contrast, in \toolName{}, gist tokens directly serve as the selection interface, allowing decoding to operate over the gist cache and thereby reducing KV-cache transfer overhead.

\paragraph{Grouped Unfolding for GQA.}
In modern architectures employing grouped-query attention (GQA)~\citep{ainslie2023gqa}, multiple query heads share the same key-value cache. Since different heads within the same KV group may prefer different chunks, we perform top-$k$ selection separately for each query head and then take the union of the selected chunk indices within the group. Let $\mathcal{H}(u)$ denote the set of query heads associated with KV group $u$. For each head $h \in \mathcal{H}(u)$, we compute $\mathcal{I}_t^{(h)} = \operatorname{Top\text{-}k}\bigl(\{s_{t,m}^{(h)}\}_{m=1}^{M}\bigr)$
and define the group-level unfolded set as: $\mathcal{I}_t^{(u)} = \bigcup_{h \in \mathcal{H}(u)} \mathcal{I}_t^{(h)}$.

All chunks in $\mathcal{I}_t^{(u)}$ are then unfolded into the shared KV-cache for group $u$. This design preserves head-specific routing while maintaining consistency with the GQA sharing pattern, ensuring that the unfolded KV-cache can be loaded once and reused across all heads in the group. This is analogous to NSA's group-centric kernel design~\citep{yuan2025native}, which loads shared KV blocks once per GQA group to maximize memory reuse and data locality. Additionally, by batching queries within each group and reusing KV blocks, it increases arithmetic intensity and enables more efficient utilization of Tensor Core matrix operations.

\subsection{Training}
\label{sec:training}
\toolName{} is trained with standard teacher forcing under the autoregressive language modeling objective, using the vanilla cross-entropy loss\footnote{Some related works instead use a KL-divergence loss against a teacher, which requires an additional forward pass with the full KV cache. This is expensive, and it caps the student's performance at the teacher's long-context ability, since the teacher's predictions serve as the upper bound.}. The training consists of two stages: continued pretraining, which is required, and selective finetuning, which is optional. Both stages share the same sequence structure: each training sequence is split into two contiguous regions---a \emph{compressed context} (the prefix) and a \emph{generation context} (the suffix). The loss is computed only over the suffix. Both stages are implemented through attention mask design, with no separate prefill or decoding phases.

\paragraph{Stage 1: Continued Pretraining (Required).}
In the continued pretraining stage, the model learns to compress context into gist tokens. The compressed context (the prefix) is augmented with interleaved gist tokens and processed under the standard gist causal mask $\mathbf{M}_{\text{gist}}$ described in \S\ref{sec:prelim}. The generation context tokens (the suffix) attend to the gist tokens produced by the compressed context, and to each other under the standard causal mask. The next token prediction loss over the generation context encourages the gist tokens to encode sufficient information for accurate next-token prediction. Since the entire forward pass is a single teacher-forced computation with a designed causal mask, training is fully parallelizable and requires no specialized kernels.
 
Importantly, after this stage alone, the model already supports selective unfolding at inference time. Because the gist tokens have learned to encode meaningful chunk-level summaries, they can be used directly as routing signals via \eqref{eq:score}--\eqref{eq:topk} without any further training. As shown in \S\ref{sec:experiments}, this inference-time selective unfolding already yields substantial gains over standard gist compression (both appended and interleaved) as well as other inference-time-only sparse attention approaches.

\paragraph{Stage 2: Selective Finetuning (Optional).}
While continued pretraining alone enables effective inference-time selection, exposing the model to the selective unfolding mechanism during training can further improve performance. In the finetuning stage, the sequence is again split into compressed and generation contexts, but the attention mask for the generation context is now modified to incorporate top-$k$ selection. Concretely: \textbf{Compressed context:} Processed identically to continued pretraining, using the standard interleaved gist mask $\mathbf{M}_{\text{gist}}$. This region produces the gist token representations.
\textbf{Generation context:} For each query position $t$ in the generation context, we compute relevance scores against all gist tokens via \eqref{eq:score} and select the top-$k$ chunks via \eqref{eq:topk}. The attention mask for position $t$ is then set to \texttt{true} only for the gist tokens and raw tokens of the selected chunks; all other compressed-context tokens are masked out.
 
Unlike continued pretraining, which employs a fixed, static attention mask, selective finetuning requires a position-dependent sparse mask where each query attends to a different subset of keys. Implementing this efficiently may require customized CUDA kernels. However, the sparse blockwise attention patterns involved are structurally similar to those in NSA~\citep{yuan2025native} and MoBA~\citep{lu2025moba}, and can leverage the same kernel design principles. We emphasize that this stage is optional: practitioners who wish to avoid the additional implementation effort can skip selective finetuning entirely and still benefit from inference-time selective unfolding after continued pretraining alone.

\subsection{Extension to Hierarchical Gist-of-Gist Compression}
\label{sec:hierarchical}
The single-level framework described above achieves linear per-step complexity of $O(M + kL)$ where $M = \lceil n/L \rceil$. We now show that the same principle---compressing representations and using them as routing signals---can be applied recursively, yielding a hierarchical structure with log-linear complexity. We refer to this extension as H-\toolName{}.

The motivation is again grounded in how humans organize and retrieve information. Long documents are rarely flat sequences---they are structured hierarchically: a book is divided into chapters, chapters into paragraphs. When searching for a specific detail, a reader does not scan every sentence from the beginning. Instead, retrieval proceeds hierarchically: one first recalls a high-level memory (e.g., the chapter or topic), then refines it to a more specific section, and finally reconstructs the precise detail. Each level acts as a progressively sharper filter over memory. H-\toolName{} mirrors this structure: meta-gist tokens encode coarse summaries, gist tokens represent finer summaries, and raw tokens preserve exact details. Hierarchical selection then implements this coarse-to-fine retrieval process, progressively narrowing attention from abstract memory to precise token-level information. We define the meta-gist tokens as follows:
\paragraph{Meta-Gist Tokens.}
Just as gist tokens summarize chunks of raw tokens, \emph{meta-gist} tokens can be defined to summarize groups of gist tokens. We introduce a meta-gist token $G_k$ after every $J$ consecutive gist-chunk pairs. The meta-gist $G_k$ summarizes the segment $\mathcal{S}_k = \{(C_j, g_j)\}_{j=(k-1)J+1}^{kJ}$, yielding $\lceil M/J \rceil$ meta-gist tokens in total. The augmented sequence becomes:
\begin{equation}
    \tilde{\mathbf{X}} = \bigl(\underbrace{C_1, g_1, \dots, C_J,  g_J}_{\text{Segment } 1},\; G_1, \;\dots,\; C_M, g_M,\; G_{\lceil M/J \rceil}\bigr).
\end{equation}
Figure~\ref{fig:overview} illustrates the hierarchical attention pattern for a two-level gist hierarchy: tokens following a meta-gist $G_k$ cannot attend to the raw tokens or gist tokens within the scope of $G_k$; they can only attend to $G_k$ itself. This blocking mechanism forces long-range history to be accessed exclusively through the meta-gist representations. The construction generalizes to arbitrary depth $t$ by introducing additional levels of summary tokens, each summarizing $J$ groups from the level below. {This means \toolName{} can theoretically support unbounded context lengths}: as the sequence grows, additional hierarchy levels are added to maintain a fixed memory footprint at the top level, with the full detail always recoverable through selective unfolding. Throughout this paper, we use the term \emph{summary tokens} to refer collectively to gist tokens and meta-gist tokens at all levels of the hierarchy, distinguishing them from the raw tokens of the original sequence.

\paragraph{Coarse-to-Fine Selection.}
{The hierarchical structure enables a coarse-to-fine selection strategy that mirrors the single-level mechanism at each scale. Instead of computing relevance scores against all $M$ gist tokens, the model first scores the $\lceil M/J  \rceil$ meta-gist tokens, selects the top-$k_2$ segments, and then scores only the $k_2 \cdot J$ gist tokens within those segments to identify the final top-$k_1$ chunks for unfolding. This reduces the per-query routing cost from $O(M)$ to $O(M/J + k_2 \cdot J)$. For a hierarchy of depth $t$, the per-query routing cost generalizes to $O(t \cdot M^{1/t})$. When $t = O(\log_J M)$, this reduces to $O(J \cdot \log_J M) = O(\log M)$---that is, each query's selection cost is \emph{logarithmic} in the context length. After selection, the query attends to the selected tokens at \emph{every} level of the hierarchy: the $k_t$ selected top-level summary tokens, the $k_{t-1}$ selected tokens at the next level, down to the $k_1$ selected gist tokens and their $k_1 \cdot L$ unfolded raw tokens. The total attention context thus contains $\sum_{\ell=1}^{t} k_\ell + k_1 \cdot L$ tokens. Since each $k_\ell$ is a small constant and $t = O(\log_J M)$, the summary tokens contribute $O(\log M)$ and the raw tokens contribute $O(k_1 L)$. The per-query attention cost is therefore $O(k_1 L + \log M)$, which matches the routing cost in order. The total per-query decoding cost is $O(\log M) = O(\log n)$, logarithmic in the context length. Over the full sequence, the aggregate decoding cost is $O(n \log n)$, which is log-linear.}

Table~\ref{tab:complexity} summarizes the complexity comparison. We provide a more detailed analysis in Appendix~\S\ref{sec:complexity}.
\begin{table}[h]
\centering
\caption{Complexity comparison. $n$: context length, $L$: chunk size, $k$: number of selected chunks, $J$: meta-gist grouping factor. All quantities treat $d$ (head dimension) as a constant. For the hierarchical variant, the depth is set to $t = \lceil \log_J M \rceil$ where $M = n/L$.}
\small
\begin{tabular}{lcc}
\toprule
\textbf{Method} & \textbf{Per-step Decoding} & \textbf{Prefill} \\
\midrule
Full Attention & $O(n)$ & $O(n^2)$ \\
\toolName{} (single-level) & $O(n/L + kL)$ & $O(nL + n^2/L^2)$ \\
\toolName{} (hierarchical) & $O(\log n + kL)$ & $O(nL + nJ/L)$ \\
\bottomrule
\end{tabular}
\label{tab:complexity}
\end{table}

\subsection{Kernel Design}
\label{sec:kernel}
 
\toolName{} replaces a dense attention operator with a sparse one, but sparsity only pays off if the kernel makes it {physical}: skipped tokens must translate into skipped FLOPs in prefill and skipped memory traffic in decoding. These two phases have opposite cost structures---prefill is compute-bound ($O(n^2)$ over the full query block), while decoding is bound by memory bandwidth and kernel-launch overhead ($q_{\text{len}}{=}1$ per step)---so we design a separate kernel for each. Both kernels compute exactly the attention defined in \S3.1--\S3.3 (no approximation beyond floating-point accumulation order); we describe them for the two-level hierarchy (H-\toolName{}), of which single-level \toolName{} is the special case with one summary level.
 
\paragraph{Prefill: block-sparse attention via key-column permutation.}
Under the gist causal mask $\mathbf{M}_{\text{gist}}$, each token attends to its local chunk, an attention sink, and all preceding summary tokens. The difficulty is that this sparsity is unstructured: summary columns recur every $L$ tokens throughout the sequence, so under the $128{\times}128$ tiling of a standard block-sparse kernel, {every} key block may contain summary columns visible to all later queries, no block is empty, and nothing is skipped. We restore structure with a key-column permutation. Before attention, we measure each key column's visibility frequency across queries and label high-frequency columns (summary tokens and the sink, visible to most queries) as {global}; we then permute the key/value sequence so all global columns come first, followed by the local columns. Because softmax normalizes over the key axis, attention is invariant under any permutation of keys, so this is exact. On the permuted layout, the same mask becomes block-structured: a thin dense slab of global columns ($q_{\text{len}} \times O(n/L_{\text{eff}})$) plus a diagonal band for the local windows, with everything else empty. We materialize this as a functional block mask and dispatch block-sparse FlexAttention, which skips the empty blocks. The permutation and block mask depend only on the mask, not on layer weights, so they are computed once per prefill and shared across all layers; only the per-layer K/V gather differs.
 
\paragraph{Decoding: selection bookkeeping without mask materialization.}
At each decoding step, the hybrid context of \eqref{eq:hybrid_kv} is fully determined by a small set of $O(n)$ metadata arrays---per-level chunk identities, gist/meta-gist indicator flags, a compressed-region flag, and the always-attended suffix---together with the per-group selection bitmaps of size $[\,\text{\#KV groups}, M{+}1\,]$ produced by grouped unfolding (\S3.3). A na\"ive implementation expands these into a dense $[H, n]$ attention mask every layer and runs masked SDPA over the {full} KV cache, which forfeits the entire benefit of sparsity: the bytes are still moved. Our decode kernel instead evaluates the keep predicate {on the fly} inside the kernel---a key $i$ in KV group $u$ is kept iff it is a raw token of a selected chunk in the compressed region, a selected gist or meta-gist token, or part of the uncompressed suffix---reading only the byte-sized metadata arrays. These arrays are query-independent within a step, so they are computed once and reused by all layers.
 
\paragraph{Decoding: three-kernel split-$K$ sparse flash-decode.}
The decode operator is factorized into three phases, each parallelized independently. \emph{(i) Parallel compaction.} A first kernel scans the keep predicate and writes the indices of kept keys into a compact per-group list. Compaction is an $O(n)$ scan, so its parallelization---not its arithmetic---determines its cost: we grid it over (\#KV groups $\times$ up to 512 blocks) and use warp-aggregated atomics (one ballot/population-count and a single \texttt{atomicAdd} per warp rather than per key) to claim output slots; output order is arbitrary, which is sound because attention is order-independent over the key set. \emph{(ii) Split-$K$ partial attention.} A second kernel computes online-softmax partial results $(m, \ell, \mathrm{acc})$ over the compact list, gridded over (KV group $\times$ split); each block derives its index range from the on-device compaction count, avoiding any host--device synchronization. Within a block, tiles of 64 kept keys' K/V are staged into shared memory \emph{once per KV group} and reused by all query heads in the group---the kernel-level counterpart of grouped unfolding (\S3.3) and analogous to NSA's group-centric design---with fp32 accumulation throughout. \emph{(iii) Combine.} A final kernel merges the per-split partials into the output with the standard log-sum-exp correction, one block per head. Decoding therefore touches only the selected $\sim k(1{+}L)$ keys plus $O(n)$ bytes of metadata, never the full KV cache.
 
 
\paragraph{End-to-end decode overhead.}

After the compact decode operator is reduced to microsecond scale, the full decoding step is dominated by host-side launch overhead from the routing path:
gist scoring, masking, top-$k$, and bitmap construction are small operations but are repeated across layers. We therefore cache all sequence-derived metadata once per decode step and share the one-token gist-state extension across layers. As an optional serving optimization, we also compile the two-level per-group selection into a graph-captured region. To make graph capture robust under variable context lengths, we bucket the number of summary tokens to powers of two and mask padded entries to $-\infty$. We may also round the selected budget up to a small grid, which selects a superset of the default top-$k$ chunks; this path is used only as an optimized serving mode, while the default compact kernel
preserves the reference selected set.
\section{Experiments}\label{sec:experiments}

\subsection{Experiment Setup}
We evaluate \toolName{} on two model families of different scales: Qwen2-7B-Instruct~\citep{yang2024qwen2technicalreport}\footnote{We choose Qwen2 because ActivationBeacon---an important baseline---has been evaluated on it.} and Llama3.2-1B~\citep{meta2024llama32blog}\footnote{We choose Llama3.2-1B because KVLink---an important baseline---has been evaluated on it.}. This allows us to assess the generality of our approach across both a large instruction-tuned model and a smaller base model.

Following the two-stage procedure described in \S\ref{sec:training}, we first perform continued pretraining with interleaved gist tokens to learn context compression, then optionally apply selective finetuning to incorporate the unfolding mechanism into training. Following ActivationBeacon~\citep{zhang2024long}, we use data sampled from RedPajama~\citep{weber2024redpajama} for continued pretraining of Qwen2-7B-Instruct, with sequence lengths ranging from 7K to 20K. Following KVLink~\citep{yang2025kvlink}, we use data sampled from FineWeb~\citep{penedo2024the} for pretraining of Llama-3.2-1B, with sequence lengths ranging from 2K to 4K. For finetuning data, we use the dataset provided by ActivationBeacon~\citep{zhang2024long} for Qwen2-7B-Instruct, which contains samples from LongAlpaca~\citep{chen2023longlora}, BookSum~\citep{kryscinski2022booksum}, and a synthesized QA dataset. For Llama-3.2-1B, we use the dataset provided by KVLink~\citep{yang2025kvlink}, which contains samples from the training sets of TriviaQA~\citep{joshi2017triviaqa} and 2WikiMQA~\citep{ho2020constructing}, with answers generated by DeepSeek-V3.2~\citep{liu2025deepseek}. All training is conducted on 8 NVIDIA H100 GPUs. Table~\ref{tab:hparams} summarizes the hyperparameters for each model and training stage.
\begin{table}[!h]
\centering
\caption{Training hyperparameters for each model and stage.}
\label{tab:hparams}
\scalebox{0.87}{
\begin{tabular}{llccccc}
\toprule
\textbf{Model} & \textbf{Stage} & \textbf{Batch Size} & \textbf{Learning Rate} & \textbf{Weight Decay} & \textbf{LR Schedule} & \textbf{\#Epoch} \\
\midrule
\multirow{2}{*}{Qwen2-7B-Inst.} & Pretraining & 8 & 1e-5 & \checkmark & Cosine w/ warmup & 1 \\
& Finetuning & 8 & 5e-6 & \checkmark & Cosine w/ warmup & 1 \\
\midrule
\multirow{2}{*}{Llama3.2-1B} & Pretraining & 128 & 1e-5 & \checkmark & Cosine w/ warmup & 1 \\
& Finetuning & 64 & 5e-6 & \checkmark & Cosine w/ warmup & 2 \\
\bottomrule
\end{tabular}
}
\end{table}

We compare against three categories of methods:
\textbf{(1) Inference-time sparse attention:} LongLLMLingua~\citep{jiang2024longllmlinguaacceleratingenhancingllms}, H$_2$O~\citep{zhang2023h2o}, StreamingLLM~\citep{xiao2023efficient}, and Quest~\citep{tang2024quest}. These methods apply KV-cache eviction or selection to models pretrained with full attention, without any additional training.
\textbf{(2) Gist-based compression:} ActivationBeacon~\citep{zhang2024long} (interleaved gist), UniGist~\citep{deng2025unigist} (interleaved gist), and KVLink~\citep{yang2025kvlink} (appended gist). These methods require continued pretraining and/or finetuning.
\textbf{(3) Full attention references:} We include the original pretrained model (no compression), continued pretraining with full attention (Full-PT), and finetuning with full attention (Full-FT) as the references. We do not include comparisons with NSA, DSA, or MoBA, as these approaches are not directly compatible with our setting: NSA requires architectural modifications, DSA depends on an external indexer, and MoBA is limited to prefill-only scenarios~\citep{lu2025moba}. Consequently, they are not applicable to the continued pretraining and sparse decoding framework we consider.

We evaluate our method under two primary settings:
(1)~\textbf{Long-context compression} on LongBench~\citep{bai2024longbench}; and (2)~\textbf{Retrieval-Augmented Generation (RAG)} on five multi-document QA benchmarks, along with a \textbf{KV-cache reuse} task. Following NSA~\citep{yuan2025native}, we exclude a subset of LongBench tasks that consistently yield low scores across models, as they offer limited discriminatory power.

The key distinction between these settings lies in how context is structured. The long-context setting treats each sample as a single continuous sequence. In contrast, in the RAG setting, each sample consists of multiple documents with explicit boundaries, and gist tokens are inserted independently within each document. 

Beyond these evaluations, we conduct extensive ablation studies to analyze the design of the hybrid context, including comparisons between adaptive top-$k$ selection and top-$p$ thresholding. We further assess \toolName{}’s fundamental retrieval capability through a passkey retrieval task across varying context lengths (Appendix~\S\ref{sec:passkey}).

\subsection{Long-context Results}
\label{sec:longbench}
Table~\ref{tab:qwen_long} reports results on LongBench using Qwen2-7B-Instruct across three compression ratios ($8\times$, $16\times$, $32\times$), under both continued pretraining and finetuning. We omit H-\toolName{} at $8\times$ compression, as the chunk size is already small. For H-\toolName{} at $16\times$, we set $L = 4$ and $J = 4$---inserting a gist token every 4 tokens and a meta-gist token every 4 gist tokens. At $32\times$, we use $L = 8$ and $J = 4$---inserting a gist token every 8 tokens.
\paragraph{\toolName{} consistently outperforms gist baselines.}
Under continued pretraining, \toolName{} achieves an average of 46.20 across tasks at $8\times$ compression, outperforming ActivationBeacon (42.52) and UniGist (43.40) by 3.7 and 2.8 points respectively, and nearly matching the Full-PT model (47.78). The advantage is maintained across higher compression ratios: at $16\times$, \toolName{} scores 45.39 versus 40.64 (ActivationBeacon) and 40.89 (UniGist); at $32\times$, \toolName{} achieves an average of 44.07 versus 38.30 and 38.16. This demonstrates that selective unfolding recovers critical fine-grained information that standard compression discards uniformly.

\paragraph{Selective finetuning further improves performance.}
With finetuning, \toolName{} continues to lead across all settings. At $8\times$, it achieves 46.74 (vs.\ 42.67 for ActivationBeacon and 43.33 for UniGist), approaching the Full-FT model (47.32). Notably, on several individual tasks \toolName{} surpasses the full-attention baseline: for instance, on MF-en task at $8\times$ finetuning, \toolName{} scores 54.24 compared to Full-FT model's 50.33, suggesting that the selective unfolding mechanism can act as a beneficial inductive bias.

\paragraph{Hierarchical variant excels at high compression.}
H-\toolName{} demonstrates its advantage at higher compression ratios where the single-level variant's routing cost becomes more significant. At $16\times$ with finetuning, H-\toolName{} achieves 46.48, surpassing \toolName{}'s 45.26. At $32\times$ with finetuning, the gap widens: H-\toolName{} scores 44.94 versus \toolName{}'s 43.35. This validates that hierarchical gist-of-gist compression and coarse-to-fine selection become more valuable as the compression ratio grows.
\begin{table}[!ht]
\centering
\caption{Performance comparison on LongBench under different compression ratios ($8\times$, $16\times$, $32\times$) using Qwen2-7B-Instruct. Best results are in bold, second-best are underlined.}
\footnotesize
\setlength{\tabcolsep}{2pt}
\scalebox{0.87}{
\begin{tabular}{c|c|ccc|ccc|cc|c|c}
\toprule
& \multirow{2}{*}{{Methods}} 
& \multicolumn{3}{c|}{{Single-Document QA}} 
& \multicolumn{3}{c|}{{Multi-Document QA}} 
& \multicolumn{2}{c|}{{Few-shot}}
& \multicolumn{1}{c|}{{Summarization}} 
& \multirow{2}{*}{{Avg.}} \\
\cmidrule{3-11}
& & NrtvQA & Qasper & MF-en 
& HotpotQA & 2WikiMQA & Musique 
& SAMSum & TriviaQA 
& GovRep & \\
\midrule
\midrule

\rowcolor{colorFULL} \cellcolor{white} -- & {Qwen2-7B-Inst.} 
& 26.33 & 42.30 & 44.50 
& 54.39 & 55.41 & 36.19 
& 45.19 & 89.91 & 36.16 
& 47.82 \\
\midrule
\multicolumn{12}{c}{{Inference-time Only}} \\
\midrule 
\rowcolor{colorX8} \cellcolor{white} & {LongLLML.}
& 15.10 & 20.21 & 22.16
& 23.74 & 28.81 & 9.94
& 39.77 & 85.98 & 20.51 
& 29.58\\
\rowcolor{colorX8} \cellcolor{white} & {H$_2$O} 
& 23.17 & 37.99 & 40.48 
& 50.66 & 50.60 & 32.05 
& 41.33 & 88.49 & 32.99 
& 44.20 \\
\rowcolor{colorX8} \cellcolor{white} & {StreamLLM} 
& 22.95 & 29.16 & 25.31 
& 44.78 & 44.56 & 27.37 
& 42.84 & 88.99 & 23.30 
& 38.81 \\
\rowcolor{colorX8} \multirow{-4}{*}{\cellcolor{white} x8} & {QUEST}
& 11.56 & 13.35 & 24.13
& 10.89 & 15.06 & 11.06
& 19.34 & 36.91 & 13.56 
& 17.32 \\
\midrule
\multicolumn{12}{c}{{Continued Pretraining}} \\
\midrule
\rowcolor{colorFULL} \cellcolor{white} -- & {Full-PT} 
& 26.59 & 38.37 & 46.39
& 57.56 & 56.28 & 36.12
& 45.83 & 89.06 & 33.78 
& 47.78 \\
\midrule
\rowcolor{colorX8} \cellcolor{white} & {ActBeacon} 
& 18.23 & \underline{35.40} & 31.69
& 48.87 & \textbf{54.45} & 30.89
& \textbf{45.30} & 87.59 & \underline{30.24} 
& 42.52 \\
\rowcolor{colorX8} \cellcolor{white} & {UniGist} 
& \underline{22.88} & 32.92 & \underline{36.52}
& \underline{55.14} & 51.10 & \underline{31.73}
& 43.02 & \underline{89.88} & 27.45 
& \underline{43.40} \\
\cmidrule{2-12}
\rowcolor{colorX8} \cellcolor{white} & {\toolName} 
& \textbf{23.55} & \textbf{36.84} & \textbf{42.21}
& \textbf{61.02} & \underline{54.23} & \textbf{32.47}
& \underline{44.59} & \textbf{90.56} & \textbf{30.36} 
& \textbf{46.20} \\
\rowcolor{colorX8} \multirow{-4}{*}{\cellcolor{white} x8}  & {H-\toolName} 
& -- & -- & -- 
& -- & -- & --
& -- & -- & --
& --  \\
\midrule
\rowcolor{colorX16} \cellcolor{white} & {ActBeacon} 
& 19.08 & \underline{32.45} & 28.57
& 46.47 & {47.93} & \underline{31.67}
& \textbf{45.50} & 85.02 & 29.06 
& 40.64 \\
\rowcolor{colorX16} \cellcolor{white} & {UniGist} 
& 21.68 & 28.90 & {35.57}
& 51.97 & 46.95 & 27.55
& 43.56 & 89.01 & 22.80 
& 40.89 \\
\cmidrule{2-12}
\rowcolor{colorX16} \cellcolor{white} & {\toolName} 
& \textbf{24.30} & \textbf{33.86} & \textbf{43.76}
& \underline{55.70} & \textbf{51.46} & \textbf{34.17}
& 44.32 & \textbf{91.02} & \underline{29.94} 
& \textbf{45.39} \\
\rowcolor{colorX16} \multirow{-4}{*}{\cellcolor{white} x16} & {H-\toolName}
& \underline{21.89} & 31.11 & \underline{37.46}
& \textbf{56.01} & \underline{50.98} & 30.25
& \underline{44.63} & \underline{89.41} & \textbf{31.05} 
& \underline{43.64} \\
\midrule
\rowcolor{colorX32} \cellcolor{white} & {ActBeacon} 
& \underline{19.90} & 27.92 & 24.71
& 43.81 & 46.95 & 28.13
& \textbf{44.70} & 83.16 & 25.42 
& {38.30} \\
\rowcolor{colorX32} \cellcolor{white} & {UniGist} 
& 18.68 & 28.89 & 26.19
& 48.67 & 43.21 & 25.73
& 43.43 & 88.49 & 20.17 
& 38.16 \\
\cmidrule{2-12}
\rowcolor{colorX32} \cellcolor{white} & {\toolName} 
& \textbf{22.61} & \textbf{34.03} & \underline{37.08}
& \textbf{54.19} & \textbf{52.03} & \underline{30.53}
& \underline{44.42} & \textbf{90.40} & \textbf{31.31} 
& \textbf{44.07} \\
\rowcolor{colorX32} \multirow{-4}{*}{\cellcolor{white} x32} & {H-\toolName} 
& 16.16 & \underline{30.82} & \textbf{39.30}
& \underline{53.66} & \underline{49.08} & \textbf{30.62}
& 44.35 & \underline{89.17} & \underline{29.89} 
& \underline{42.56} \\
\midrule
\multicolumn{12}{c}{{Finetuning}} \\
\midrule
\rowcolor{colorFULL} \cellcolor{white} -- & {Full-FT} 
& 27.84 & 38.56 & 50.33
& 52.26 & 54.87 & 35.25
& 43.70 & 87.99 & 35.07 
& 47.32 \\
\midrule
\rowcolor{colorX8} \cellcolor{white}
& {ActBeacon} 
& \textbf{26.85} & 36.95 & \underline{47.93} 
& 48.91 & \underline{41.27} & 22.54 
& \textbf{44.61} & 87.32 & \underline{27.62} 
& 42.67 \\
\rowcolor{colorX8} \cellcolor{white} & {UniGist} 
& 24.17 & \underline{38.27} & 47.28
& \underline{54.02} & {41.26} & \underline{24.84}
& 43.63 & \underline{89.92} & 26.59 
& \underline{43.33} \\
\cmidrule{2-12}
\rowcolor{colorX8} \cellcolor{white} & {\toolName} 
& \underline{26.24} & \textbf{43.54} & \textbf{54.24}
& \textbf{54.26} & \textbf{48.18} & \textbf{28.95}
& \underline{44.31} & \textbf{90.11} & \textbf{30.79} 
& \textbf{46.74} \\
\rowcolor{colorX8} \multirow{-4}{*}{\cellcolor{white} x8}  & {H-\toolName} 
& -- & -- & -- 
& -- & -- & --
& -- & -- & --
& -- \\
\midrule
\rowcolor{colorX16} \cellcolor{white} 
& {ActBeacon} 
& \underline{23.81} & 35.77 & 38.91
& 44.29 & 41.69 & 22.76
& \textbf{46.39} & 86.06 & 27.80 
& {40.83} \\
\rowcolor{colorX16} \cellcolor{white} & {UniGist} 
& 23.15 & 34.91 & {40.54}
& 46.79 & 38.67 & 22.82
& \underline{43.82} & \underline{89.97} & 23.15 
& 40.42 \\
\cmidrule{2-12}
\rowcolor{colorX16} \cellcolor{white} & {\toolName} 
& \textbf{24.38} & \underline{42.47} & \underline{51.98}
& \underline{54.69} & \underline{42.91} & \underline{28.59}
& 42.66 & 88.43 & \textbf{31.27} 
& \underline{45.26} \\
\rowcolor{colorX16} \multirow{-4}{*}{\cellcolor{white} x16} & {H-\toolName} 
& 22.72 & \textbf{43.98} & \textbf{53.40}
& \textbf{56.11} & \textbf{48.31} & \textbf{29.04}
& 43.28 & \textbf{90.36} & \underline{31.12} 
& \textbf{46.48} \\
\midrule
\rowcolor{colorX32} \cellcolor{white} & {ActBeacon} 
& {21.32} & {32.52} & 33.18
& {41.05} & \underline{42.44} & 19.10
& \underline{43.04} & 85.66 & {28.33} 
& {38.52} \\
\rowcolor{colorX32} \cellcolor{white} & {UniGist} 
& 19.96 & 31.98 & {33.87}
& 40.34 & 37.63 & {20.47}
& \textbf{43.43} & \textbf{89.56} & 21.32 
& 37.62 \\
\cmidrule{2-12}
\rowcolor{colorX32} \cellcolor{white} & {\toolName} 
& \textbf{24.58} & \underline{40.24} & \underline{45.68}
& \underline{50.64} & 39.38 & \underline{27.55}
& {42.69} & 89.10 & \underline{30.30} 
& \underline{43.35} \\
\rowcolor{colorX32} \multirow{-4}{*}{\cellcolor{white} x32} & {H-\toolName} 
& \underline{22.03} & \textbf{43.32} & \textbf{50.79}
& \textbf{52.47} & \textbf{45.18} & \textbf{27.93}
& 42.56 & \underline{89.47} & \textbf{30.68} 
& \textbf{44.94} \\
\bottomrule
\end{tabular}
}
\label{tab:qwen_long}
\end{table}

\subsection{RAG Results}
\label{sec:rag}
Table~\ref{tab:main_results} reports results on five multi-document QA benchmarks using Llama-3.2-1B across two compression ratios ($8\times$, $16\times$), under both continued pretraining and finetuning.

The RAG setting reveals the most striking advantage of \toolName{}. At $8\times$ compression with continued pretraining only, \toolName{} achieves 33.68 average score, not only outperforms KVLink (21.58) and UniGist (22.53) by over 11 points, but \textbf{even {surpasses} the vanilla (27.99) and Full-PT model (27.14).} We attribute this to the nature of RAG: each query is relevant to only one of the retrieved documents, while the remaining documents act as distractors. Full attention distributes capacity uniformly across all documents, diluting focus on the relevant passage. In contrast, \toolName{}'s selective unfolding mechanism concentrates attention on the most query-relevant chunks, effectively filtering out distracting documents.

After finetuning, \toolName{} at $8\times$ achieves 53.39, approaching Full-FT (57.76). This represents an improvement of 11.94 points over KVLink (41.45) and 8.12 points over UniGist (45.27). At $16\times$, \toolName{} scores 49.29, outperforming KVLink's 38.67 and UniGist's 42.24.
 
Consistent with the long-context results, H-\toolName{} outperforms the single-level variant at $16\times$ compression. Under continued pretraining, H-\toolName{} achieves 30.19, compared to 27.74 for \toolName{}. After finetuning, H-\toolName{} reaches 50.72 versus 49.29 for \toolName{}, suggesting that the coarse-to-fine hierarchical selection more effectively allocates the limited token budget.

A notable pattern in the RAG results is that Full-PT underperforms \toolName{} under continued pretraining, yet surpasses it after finetuning. This is because continued pretraining optimizes a generic next-token prediction objective over the given context, without any notion of a query or a target answer. The model has no signal to distinguish relevant from irrelevant documents, and thus learns to attend to all content indiscriminately. Finetuning, by contrast, trains the model on the downstream task format---given multiple documents and a question, produce an answer---which explicitly teaches it to focus on query-relevant passages and suppress distractors.
\begin{table}[!ht]
\centering
\caption{Performance comparison on RAG under different compression ratios ($8\times$, $16\times$) using Llama3.2-1B. Best results are in bold, second-best are underlined.}
\label{tab:main_results}
\scalebox{0.87}{
\begin{tabular}{c|c|ccccc|c}
\toprule
& {Methods} & {HotpotQA} & {TriviaQA} & {2WikiMQA} & {MuSiQue} & {NQ} & {Avg.} \\
\midrule
\midrule
\rowcolor{colorFULL} \cellcolor{white} -- & {Llama3.2-1b} & 29.98 & 40.82 & 25.86 & 9.10  & 34.20 & 27.99 \\
\midrule
\multicolumn{8}{c}{{Continued Pretraining}}\\
\midrule
\rowcolor{colorFULL} \cellcolor{white} -- & {Full-PT }  & 29.64 & 39.73 & 25.45 & 9.18  & 31.72 & 27.14 \\
\midrule
\rowcolor{colorX8} \cellcolor{white} & {KVLink}  & 17.54 & 44.36 & 25.03 & 4.01 & 16.96 & 21.58 \\
\rowcolor{colorX8} \cellcolor{white} & {UniGist}  & 22.77 & 44.44 & 22.84 & 3.72  & 18.86 & 22.53 \\
\cmidrule{2-8}
\rowcolor{colorX8} \cellcolor{white} & {\toolName} & \textbf{41.09} & \textbf{52.06} & \textbf{38.65} & \textbf{12.99} & \textbf{23.62} & \textbf{33.68} \\
\rowcolor{colorX8} \multirow{-4}{*}{\cellcolor{white} x8} & {H-\toolName}  & -- & -- & -- & --  & -- & -- \\
\midrule

\rowcolor{colorX16} \cellcolor{white} & {KVLink}      & 10.67 & 33.10 & 21.22 & 0.74 & 4.92 & 14.13  \\
\rowcolor{colorX16} \cellcolor{white} & {UniGist}  & 21.51 & 41.62 & 30.10 & 3.23 & 14.56 & 22.20 \\
\cmidrule{2-8}

\rowcolor{colorX16} \cellcolor{white} & {\toolName} & 30.56 & 46.28 & 32.67 & 8.11 & 21.10 & 27.74 \\
\rowcolor{colorX16} \multirow{-4}{*}{\cellcolor{white} x16} & {H-\toolName}  & \textbf{34.34} & \textbf{50.98} & \textbf{33.61} & \textbf{9.06} & \textbf{22.98} & \textbf{30.19} \\
\midrule
\multicolumn{8}{c}{{Finetuning}}\\
\midrule
\rowcolor{colorFULL} \cellcolor{white} -- & {Full-FT}  & 63.82 & 71.38 & 75.83 & 24.20 & 53.56 & 57.76 \\
\midrule
\rowcolor{colorX8} \cellcolor{white} & {KVLink} & 41.15 & 61.65 & 62.68 & 10.47 & 31.28 & 41.45 \\
\rowcolor{colorX8} \cellcolor{white} & {UniGist}  & 49.25 & 64.00 & 64.17 & 12.37 & 36.56 & 45.27 \\
\cmidrule{2-8}
\rowcolor{colorX8} \cellcolor{white} & {\toolName}  & \textbf{59.88} & \textbf{68.62} & \textbf{71.99} & \textbf{20.27} & \textbf{46.20} & \textbf{53.39} \\
\rowcolor{colorX8} \multirow{-4}{*}{\cellcolor{white} x8} & {H-\toolName}    & -- & -- & -- & --  & -- & -- \\
\midrule
\rowcolor{colorX16} \cellcolor{white} & {KVLink}  & 36.92 & 58.85 & 62.21 & 8.11 & 27.24 & 38.67 \\
\rowcolor{colorX16} \cellcolor{white} & {UniGist}    & 43.50 & 61.83 & 60.72 & 10.88 & 34.26 & 42.24 \\

\cmidrule{2-8}
\rowcolor{colorX16} \cellcolor{white} & {\toolName}  & 54.29 & 66.16 & 67.93 & 16.30 & 41.78 & 49.29 \\
\rowcolor{colorX16} \multirow{-4}{*}{\cellcolor{white} x16} & {H-\toolName} & \textbf{56.43} & \textbf{66.76} & \textbf{69.56} & \textbf{18.37} &\textbf{42.48} & \textbf{50.72} \\

\bottomrule
\end{tabular}
}
\end{table}

\subsection{Parallel Computing with KV-cache Reuse}
\label{sec:kv_reuse}
KV-cache reuse---where KV caches of long documents are stored and reused across repeated queries to avoid redundant prefilling and support parallel computing---is a natural fit for \toolName{}. In this setting, \toolName{} encodes documents independently with interleaved gist tokens and reuses their KV caches at query time.

Following the KVLink~\citep{yang2025kvlink} protocol, we insert link tokens between documents. These Link tokens attend only to gist and meta-gist tokens from preceding documents, enabling efficient cross-document routing, while selective unfolding recovers full-resolution detail only for the most relevant segments. This avoids traversing full raw KV caches and improves efficiency without sacrificing quality. In Table~\ref{tab:kv-reuse}, \toolName{} achieves an average RAG performance of 48.07 at $8\times$ compression and 41.59 at $16\times$, substantially outperforming KVLink and UniGist; H-\toolName{} further improves the $16\times$ compression result to 46.34. These results support gist tokens as effective routing signals for cross-document reasoning under KV-cache reuse.

\begin{table}[!h]
\centering
\caption{RAG benchmarks with KV-cache Reuse under different compression ratios ($8\times$, $16\times$) using Llama3.2-1B. Best results are in bold.}
\label{tab:kv-reuse}
\scalebox{0.87}{
\begin{tabular}{c|c|ccccc|c}
\toprule
& {Methods} & {HotpotQA} & {TriviaQA} & {2WikiMQA} & {MuSiQue} & {NQ} & {Avg.} \\
\midrule
\rowcolor{colorX8} \cellcolor{white}  & {KVLink} & 41.55 & 60.00 & 63.91 & 9.68 & 30.86 
& 41.20 \\
\rowcolor{colorX8} \cellcolor{white}  & {UniGist}  & 42.74 & 59.98 & 61.51 & 9.02 & 28.48 
& 40.35 \\
\cmidrule{2-8}
\rowcolor{colorX8} \cellcolor{white}  & {\toolName}  & \textbf{54.23} & \textbf{64.81} & \textbf{70.21} & \textbf{13.98} & \textbf{37.10} & \textbf{48.07} \\
\rowcolor{colorX8} \multirow{-4}{*}{\cellcolor{white} x8} & {H-\toolName}    & -- & -- & -- & --  & -- & -- \\
\midrule
\rowcolor{colorX16} \cellcolor{white}  & {KVLink}  & 37.11 & 57.97 & 61.43 & 8.03 & 27.84 & 38.48 \\
\rowcolor{colorX16} \cellcolor{white}  & {UniGist} & 37.20 & 57.65 & 58.14 & 7.53 & 25.42 
& 37.19 \\

\cmidrule{2-8}
\rowcolor{colorX16} \cellcolor{white}  & {\toolName}  & {44.58} & 59.32 & 61.88 & 9.68 & 32.50 & 41.59 \\
\rowcolor{colorX16} \multirow{-4}{*}{\cellcolor{white} x16} & {H-\toolName} & \textbf{50.95} & \textbf{62.77} & \textbf{68.26} & \textbf{12.25} &\textbf{37.46} & \textbf{46.34} \\

\bottomrule
\end{tabular}
}
\end{table}

\subsection{Ablation Analysis}\label{sec:ablation}
\paragraph{Selective Rules Comparison.}
Table~\ref{tab:abl-srule} ablates the design of the hybrid attention context (Eq.~\eqref{eq:hybrid_kv}). We compare three variants: (1)~\textbf{SR-only}: the attention context includes only the selected raw tokens, without their corresponding gist tokens; (2)~\textbf{AG+SR}: all gist tokens are retained in the context alongside the selected raw tokens; and (3)~\textbf{SG+SR}: only the selected gist tokens are retained together with their raw tokens (our default). SG+SR achieves the best average score (53.39), outperforming SR-only (52.76) and slightly surpassing AG+SR (53.27). The result confirms our design choice in \S\ref{sec:hybrid}: retaining selected gist tokens alongside their raw tokens provides complementary compressed information, while including all gist tokens (AG+SR) dilutes the attention with less relevant summaries. Dropping gist tokens entirely (SR-only) loses the compressed cross-chunk patterns that gist tokens capture.
\begin{table}[!ht]
\centering
\caption{Ablation Study on different selective rules. We adopt SG+SR as the default configuration.}
\scalebox{0.87}{
\begin{tabular}{l|ccccc|c}
\toprule 
{Variants} & {HotpotQA} & {TriviaQA} & {2WikiMQA} & {MuSiQue} & {NQ} & {Avg.} \\
\midrule
SR-only
& 58.74 & 68.33 & 71.70 & \textbf{20.76} & 44.28 & 52.76 \\

AG+SR
& \underline{59.53} & \textbf{68.74} & \underline{71.72} & \underline{20.68} & \underline{45.70} & \underline{53.27} \\

SG+SR
& \textbf{59.88} & \underline{68.62} & \textbf{71.99} & 20.27 & \textbf{46.20} & \textbf{53.39} \\
\bottomrule
\end{tabular}
}
\label{tab:abl-srule}
\end{table}

\begin{figure}
    \centering
    \includegraphics[width=1.0\linewidth]{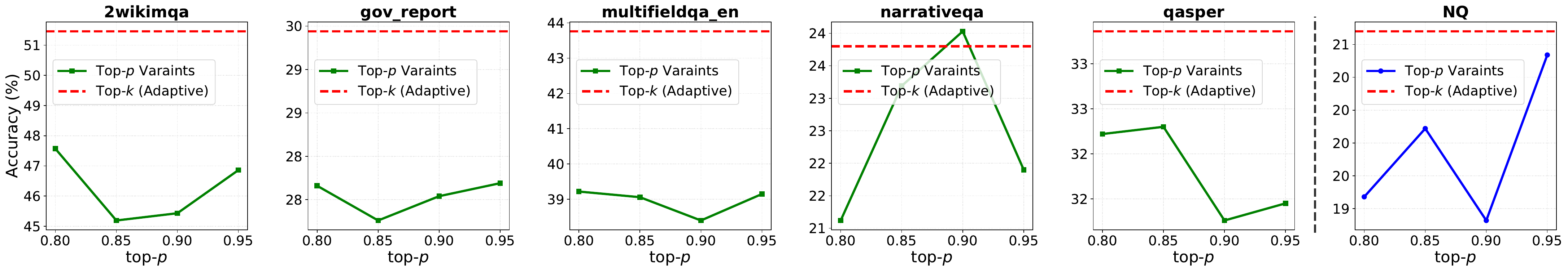}
    \caption{Top-$k$ vs.\ top-$p$ selection across five LongBench tasks (on Qwen2) and NQ (on Llama3.2). The red dashed line indicates our adaptive top-$k$ selection; green/blue solid lines show top-$p$ variants with $p \in \{0.80, 0.85, 0.90, 0.95\}$. Adaptive top-$k$ consistently outperforms all top-$p$ thresholds and exhibits greater stability across tasks.}
    \label{fig:topp_ablation}
\end{figure}

\paragraph{Top-$k$ vs. Top-$p$.}
Figure~\ref{fig:topp_ablation} compares our adaptive top-$k$ selection (red dashed line) against top-$p$ (nucleus) selection with varying thresholds $p \in \{0.80, 0.85, 0.90, 0.95\}$ (green/blue solid lines) across six LongBench tasks. Top-$p$ selection dynamically adjusts the number of unfolded chunks by including chunks until their cumulative score mass reaches $p$. Across all six tasks, adaptive top-$k$ consistently outperforms every top-$p$ threshold, often by a substantial margin. For instance, on 2WikiMQA, top-$k$ achieves over 51\% while the best top-$p$ variant ($p=0.80$) reaches only 47.5\%. On GovReport, top-$k$ scores approximately 30 compared to 28--29 for all top-$p$ settings. The performance of top-$p$ is also notably unstable across tasks: on NarrativeQA it peaks at $p=0.85$ but drops sharply at other thresholds, and on NQ it fluctuates between 19 and 21 with no consistent trend. This instability arises because the score distribution varies across queries and tasks, making a fixed probability threshold a poor proxy for the number of relevant chunks. In contrast, adaptive top-$k$ directly controls the token budget, providing a more reliable and stable selection mechanism.

\subsection{Efficiency and Scalability}
\label{sec:efficiency}
 
\paragraph{Setup.}
We benchmark the kernels of \S3.6 on a single NVIDIA H100 80GB (3.35\,TB/s HBM3) with the finetuned SSA Qwen2-7B-Instruct model at $16\times$ compression (28 query heads, 4 KV heads, $d{=}128$, bf16). We evaluate two variants: a single-level {SSA} ($L{=}16$) and the hierarchical {H-SSA} ($L{=}4$, $J{=}4$). The dense baselines are: FlashAttention~\citep{dao2022flashattention} for prefill and Flash-Decoding~\citep{dao2023flashdecoding} for decoding. We report time-to-first-token (TTFT) for prefill and time-per-output-token (TPOT) for decoding.
 
\begin{figure}[t]
\centering
\includegraphics[width=\linewidth]{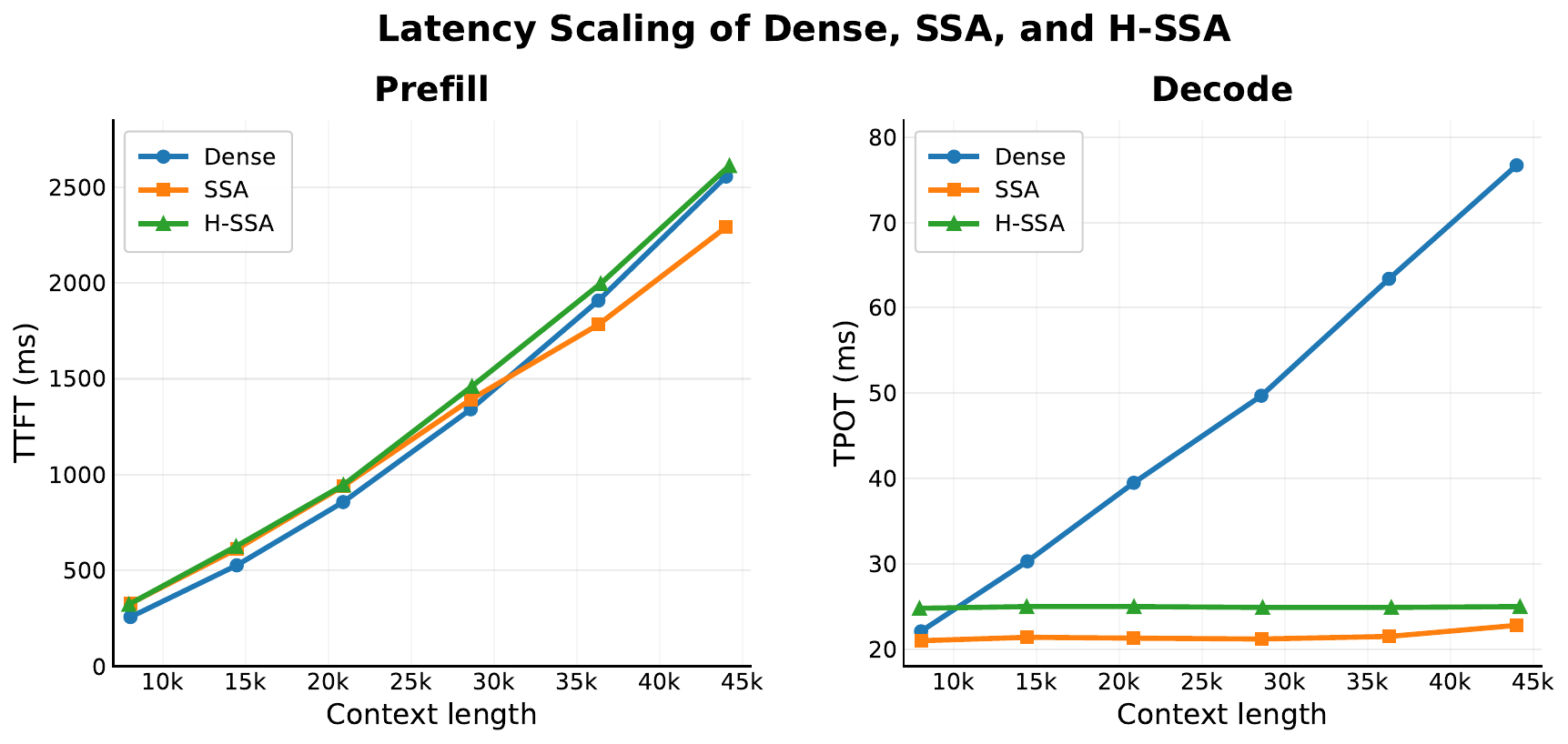}
\caption{\textbf{End-to-end latency scaling on Qwen2-7B.} During prefill, SSA and H-SSA achieve TTFT comparable to the dense FlashAttention baseline across the measured context range (8K--44K). During decoding, the main scalability benefit emerges: dense Flash-Decoding's TPOT grows approximately linearly with context length, whereas SSA and H-SSA remain nearly flat.}
\label{fig:latency}
\end{figure}

\paragraph{End-to-end scaling.}
As shown in Figure~\ref{fig:latency}, the clearest end-to-end gain comes from decoding. Dense Flash-Decoding rereads the entire KV cache at every step, so its per-token latency grows with context length, increasing from 21.9 ms at 8K to 76.4 ms at 44K. In contrast, our method reads only around~1\% of the cache, making this cost negligible rather than dominant. As a result, its per-token latency remains nearly flat: the two-level H-\toolName{} stays close to 25 ms, while the single-level \toolName{} remains around 21--23 ms across the full range. Consequently, the speedup widens with context length, reaching 3.05$\times$ for H-\toolName{} and 3.37$\times$ for \toolName{} at 44K. For prefill, \toolName{} keeps pace end-to-end with FlashAttention and pulls ahead at longer contexts. H-\toolName{} is 1.29$\times$ slower at 8K but closes the gap to parity by 44K. The single-level \toolName{} crosses over near 33K and becomes faster than the baseline thereafter, reaching 0.90$\times$ the dense latency at 44K.

\begin{table}[t]
\centering
\caption{
Attention-operator scaling for dense and H-\toolName{} operators on an H100 GPU.
}
\label{tab:operator_scaling}
\small
\begin{tabular}{r|ccc|ccc}
\hline
& \multicolumn{3}{c|}{Prefill operator (ms)} &
  \multicolumn{3}{c}{Decode operator ($\mu$s)} \\
$k_{\mathrm{len}}$ & FlashAttn & H-\toolName{} & speedup & Flash-Dec & H-\toolName{}  & speedup \\
\hline
32{,}768  & 22.30  & 3.82   & 5.8$\times$ & 44.2  & 35.6 & 1.24$\times$ \\
50{,}000  & 52.70  & 7.95   & 6.6$\times$ & 57.1  & 40.8 & 1.40$\times$ \\
65{,}536  & 90.13  & 13.17  & 6.8$\times$ & 64.7  & 45.5 & 1.42$\times$ \\
100{,}000 & 226.83 & 28.28  & 8.0$\times$ & 88.1  & 56.3 & 1.56$\times$ \\
131{,}072 & 393.50 & 47.71  & 8.2$\times$ & 107.7 & 66.3 & 1.62$\times$ \\
200{,}000 & 924.18 & 111.31 & {8.3}$\times$ & 151.9 & 96.3 & 1.58$\times$ \\
\hline
\end{tabular}
\end{table}

\paragraph{Attention-operator scaling.}
The operator-level results in Table~\ref{tab:operator_scaling} isolate the
attention computation from model-level and
host-level overheads; we use H-\toolName{} as the representative configuration.
For prefill, its sparse operator is consistently faster than dense
FlashAttention, with the gap widening as sequence length increases: the speedup
rises from $5.8\times$ at 32K tokens to $8.3\times$ at 200K tokens. For
decoding, the sparse operator is $1.24$--$1.62\times$ faster than
Flash-Decoding over 32K--200K tokens.

\section{Related Works}
\paragraph{Sparse Attention Mechanisms.}
Addressing the quadratic complexity of the self-attention mechanism has been a central theme in scaling Transformers to long contexts. Early approaches introduced fixed or heuristic sparsity patterns. {Sparse Transformer} \citep{child2019generating} and {Longformer} \citep{beltagy2020longformer} utilize fixed local windows combined with global attention tokens to reduce complexity to $O(N \sqrt{N})$ or $O(N)$. Many recent KV-cache management works such as {H$_2$O} \citep{zhang2023h2o}, StreamingLLM \citep{xiao2023efficient}, LongLLMLingua~\citep{jiang2024longllmlinguaacceleratingenhancingllms}, and Quest~\citep{tang2024quest} aim to evict less important KV pairs. More recently, hardware-aware and learnable sparse attention mechanisms---such as Native Sparse Attention \citep{yuan2025native}, DeepSeek Sparse Attention \citep{liu2025deepseek}, and MoBA \citep{lu2025moba}---have been proposed to dynamically identify and attend to the most relevant blocks in a hardware-efficient manner.

\paragraph{Context Compression.}
An alternative approach to efficient long-context processing is compressing the context into compact representations. Gist Tokens~\citep{mu2023learning} and AutoCompressors~\citep{chevalier2023adapting} train models to compress prompt instructions into special tokens via an attention mask bottleneck. KVLink~\citep{yang2025kvlink} appends learnable compression tokens to the end of each document, where each token can attend to the entire preceding context to produce a global summary. ActivationBeacon~\citep{zhang2024long}, UniGist~\citep{deng2025unigist}, and GistPool~\citep{petrov2025long} interleave gist tokens throughout the sequence, associating each with a local chunk. While these methods effectively reduce sequence length, they typically treat compression as a one-way process: once the context is compressed, the raw tokens are discarded and irreversibly lost. 

\paragraph{Alternative Sequence Modeling Architectures.}
A separate line of research seeks to replace the softmax attention mechanism entirely with architectures that achieve linear or near-linear complexity. MultiresLayer~\citep{shi_multires} captures multiscale trends in the input sequence using convolutional networks. Linear attention methods~\citep{katharopoulos2020transformers} approximate the softmax kernel with feature maps, enabling recurrent computation with a fixed-size state. State space models such as S4~\citep{gu2021efficiently} and Mamba~\citep{gu2023mamba} model sequences through structured recurrences with selective gating, achieving linear-time inference with competitive language modeling performance. Recurrent alternatives including RWKV~\citep{peng2023rwkv} and RetNet~\citep{sun2023retentive} combine recurrent architectures with attention-like mechanisms, offering constant-memory inference. More recent gated linear attention variants such as GLA~\citep{yang2024gated}, refine selective forgetting through per-token gates. These approaches represent a fundamentally different design axis from our work. They modify or replace the core attention mechanism with alternative sequence modeling primitives, often requiring pretraining from scratch on the new architecture.

\paragraph{Memory-Augmented Language Models.}
A complementary line of research equips language models with external memory modules to handle contexts that exceed their native attention window. LongMem~\citep{wang2023augmenting} decouples memory encoding from retrieval by maintaining a frozen backbone alongside a trainable side network that retrieves cached key--value pairs from past inputs. MemGPT~\citep{packer2023memgpt} treats the LLM as an operating system that explicitly pages information between a limited context window and an external store, enabling long-horizon reasoning over conversations and documents. More recent systems such as Mem0~\citep{chhikara2025mem0} and MemOS~\citep{li2025memos} extend this idea by introducing structured, persistent memory layers with mechanisms for adding, updating, and consolidating memories across sessions. From this perspective, \toolName{} can itself be viewed as a memory mechanism: each gist token is a learned latent summary of a local chunk that is stored in the KV-cache and later retrieved on demand via attention-based scoring, while meta-gist tokens form a hierarchical memory index that supports coarse-to-fine recall---closely paralleling the multi-tier memory hierarchies of MemGPT and MemOS.
\section{Conclusions}
\label{sec:discussion}
We presented \toolName{}, a framework that unifies gist-based context compression with sparse attention through selective unfolding. By leveraging interleaved gist tokens as both compressed summaries and learned routing signals, \toolName{} enables query-adaptive recovery of fine-grained detail without architectural modifications, external indexers, or non-differentiable gates. The framework naturally extends to hierarchical gist-of-gist compression, achieving logarithmic per-step decoding complexity through coarse-to-fine selection. Experiments on LongBench and RAG benchmarks demonstrate consistent gains over both gist compression and sparse attention baselines across compression ratios from $8\times$ to $32\times$. Latency measurements further show that \toolName{} achieves up to a
$3.37\times$ end-to-end decoding speedup over Flash-Decoding, while matching and eventually outperforming the FlashAttention baseline during prefill.

\section{Acknowledgment}
This work was supported in part by ONR Grant N00014-22-1-2110, NSF Grant 2205084, and the Stanford Institute for Human-Centered Artificial Intelligence (HAI). EBF is a Biohub, San Francisco, Investigator.
\newpage

\bibliography{bibliography}
\bibliographystyle{plainnat}

\newpage
\appendix
\onecolumn

\crefalias{section}{appsec}
\crefalias{subsection}{appsec}
\crefalias{subsubsection}{appsec}

\setcounter{equation}{0}
\renewcommand{\theequation}{\thesection.\arabic{equation}}

\onecolumn
\section*{\LARGE Supplementary Material}
\label{sec:appendix}



\section{More Details on Adaptive Top-k.}
\label{exp:topk}
Concretely, let $n_{\text{kv}}$ denote the total number of KV positions available for the generation context, $L_{\text{eff}}$ the effective compression factor (equal to the chunk size $L$ for single-level \toolName{}, or $L \cdot J$ for the hierarchical variant H-\toolName{}), and $G$ the number of query heads per KV group. The per-level selection budget is computed as:
\begin{equation}
    k = \left\lfloor \frac{n_{\text{kv}}}{L_{\text{eff}} \cdot G \cdot L} \right\rfloor + 1,
\end{equation}
where the division by $G$ accounts for the grouped unfolding mechanism (\S\ref{sec:hybrid}): since each head within a GQA group may select different chunks and their union is unfolded into the shared KV-cache, the effective number of unfolded chunks per group scales with $G$. The $+1$ ensures that at least one chunk is always selected. For the hierarchical variant, the same $k$ is used at both the meta-gist and gist levels. This adaptive scheme ensures that the total number of unfolded tokens remains approximately proportional to $n_{\text{kv}}$ regardless of the compression ratio, providing a fair comparison across settings.

\section{Complexity Analysis}
\label{sec:complexity}
 
We provide a detailed complexity analysis for both the single-level and hierarchical variants of \toolName{}. Let $n$ denote the input length, and $d$ the head dimension (treated as a constant).

\subsection{Single-Level Complexity}
Given chunk size $L$, the number of chunks (and thus gist tokens) is $M = \lceil n / L \rceil$, and let $k$ denote the number of chunks selected for unfolding.

\paragraph{Prefill.}
During prefill, the computation decomposes into two parts:
\begin{enumerate}
    \item \textbf{Intra-chunk attention.} Each chunk $C_m$ has $L$ raw tokens plus one gist token. Under the causal mask, tokens within $C_m$ attend to at most $L + 1$ positions. Across all $M$ chunks, this costs $O(M \cdot L \cdot (L+1) \cdot d) = O(nL)$, since $M = \lceil n/L  \rceil$ and $d$ is constant.
    \item \textbf{Gist-to-gist attention.} Each gist token $g_m$ can attend to all preceding gist tokens $g_1, \dots, g_{m-1}$. The total cost across all $M$ gist tokens is $\sum_{m=1}^{M} m = O(M^2) = O(n^2 / L^2)$.
\end{enumerate}
The overall prefill cost is therefore $O(nL + n^2/L^2)$. For typical settings where $L = O(\sqrt{n})$, both terms are $O(n\sqrt{n})$, yielding sub-quadratic prefill. For fixed $L$, the gist-to-gist term $O(n^2/L^2)$ dominates, which is quadratic but reduced by a factor of $L^2$ compared to full attention.

\paragraph{Decoding (per step).}
At each decoding step $t$, the computation involves:
\begin{enumerate}
    \item \textbf{Relevance scoring.} The query $\mathbf{q}_t$ computes dot products with all $M$ gist keys: $s_{t,m} = \mathbf{q}_t^\top \mathbf{k}_{g_m} / \sqrt{d}$ for $m = 1, \dots, M$. This costs $O(M \cdot d) = O(M)$.
    \item \textbf{Top-$k$ selection.} Selecting the $k$ highest scores from $M$ candidates costs $O(M)$ (via a partial sort).
    \item \textbf{Hybrid attention.} The query attends to $k$ gist tokens and $k \cdot L$ raw tokens from the selected chunks, totaling $k(1 + L)$ key-value pairs. The attention computation costs $O(k(1+L) \cdot d) = O(kL)$.
\end{enumerate}
The total per-step decoding cost is:
\begin{equation}
    \underbrace{O(M)}_{\text{scoring}} + \underbrace{O(M)}_{\text{top-}k} + \underbrace{O(kL)}_{\text{attention}} = O(M + kL) = O\!\left(n/L + kL\right).
\end{equation}
Since $k$ and $L$ are fixed constants independent of $n$, this is \emph{linear} in the context length $n$. Over a full sequence of $n$ decoding steps, the aggregate decoding cost is $O(n^2/L + nkL)$, which is quadratic in $n$ but reduced by a factor of $L$ compared to full attention's $O(n^2)$.
 
\subsection{Hierarchical Complexity}
 
We now analyze the hierarchical variant with meta-gist grouping factor $J$ and hierarchy depth $t$. At each level $\ell \in \{1, \dots, t\}$, there are $M_\ell = \lceil M / J^{\ell-1} \rceil$ summary tokens: $M_1 = M$ gist tokens at the bottom, $M_2 = \lceil M/J \rceil$ meta-gist tokens at level 2, and so on up to $M_t = \lceil M / J^{t-1} \rceil$ tokens at the top.

\paragraph{Prefill.}
The blocking mechanism ensures that tokens at level $\ell$ can only attend to summary tokens at the same or higher levels within their local scope. The computation decomposes as:
\begin{enumerate}
    \item \textbf{Intra-chunk attention.} Same as the single-level case: $O(nL)$.
    \item \textbf{Level-$\ell$ summary attention.} At level $\ell$, each of the $M_\ell$ summary tokens attends to at most $J$ tokens from level $\ell - 1$ (its children) plus the preceding summary tokens at level $\ell$. Due to the blocking mechanism, each level-$\ell$ token's effective attention span is bounded by $O(J + M_\ell)$. However, the blocking ensures that level-$\ell$ tokens only see level-$\ell$ peers within their parent's scope at level $\ell+1$, bounding the span to $O(J)$ per token. The total cost at level $\ell$ is $O(M_\ell \cdot J)$. Summing across all levels:
    \begin{equation}
        \sum_{\ell=1}^{t} O(M_\ell \cdot J) = O\!\left(J \sum_{\ell=1}^{t} \frac{M}{J^{\ell-1}}\right) = O\!\left(MJ \cdot \frac{1 - J^{-t}}{1 - J^{-1}}\right) = O(MJ).
    \end{equation}
    The geometric series converges since $J \geq 2$, so the total summary attention cost is $O(MJ) = O(nJ/L)$, independent of the depth $t$.
\end{enumerate}
The overall prefill cost is therefore $O(nL + nJ/L)$, which is \emph{linear} in $n$---a substantial improvement over the single-level prefill cost of $O(nL + n^2/L^2)$.
 
\paragraph{Decoding: Routing.}
Coarse-to-fine selection proceeds top-down through the hierarchy. At each level $\ell$, the model selects $k_\ell$ summary tokens from the candidates exposed by the previous level's selection:
\begin{enumerate}
    \item \textbf{Level $t$ (top).} Score all $M_t = \lceil M/J^{t-1} \rceil$ top-level tokens and select the top-$k_t$. Cost: $O(M_t)$.
    \item \textbf{Level $\ell < t$.} The $k_{\ell+1}$ selected tokens at level $\ell+1$ expose $k_{\ell+1} \cdot J$ candidate tokens at level $\ell$. Score these and select the top-$k_\ell$. Cost: $O(k_{\ell+1} \cdot J)$.
\end{enumerate}
The total routing cost is:
\begin{equation}
    \underbrace{O(M_t)}_{\text{top level}} + \sum_{\ell=1}^{t-1} \underbrace{O(k_{\ell+1} \cdot J)}_{\text{level } \ell} = O\!\left(\frac{M}{J^{t-1}} + (t-1) \cdot k_{\max} \cdot J\right),
\end{equation}
where $k_{\max} = \max_\ell k_\ell$. When $t = \lceil \log_J M \rceil$, the first term becomes $O(M / J^{\log_J M - 1}) = O(J)$, a constant. The second term is $O(k_{\max} \cdot J \cdot \log_J M)$. Since $k_{\max}$ and $J$ are fixed constants, the total routing cost is:
\begin{equation}
    O(J + k_{\max} J \log_J M) = O(\log M) = O(\log(n/L)) = O(\log n).
\end{equation}
 
\paragraph{Decoding: Attention.}
After routing, the query attends to the selected tokens at every level of the hierarchy: $k_\ell$ selected summary tokens at each level $\ell \in \{2, \dots, t\}$, plus $k_1$ selected gist tokens and their $k_1 \cdot L$ unfolded raw tokens at the bottom level. The total number of key-value pairs in the attention context is $\sum_{\ell=1}^{t} k_\ell + k_1 \cdot L$. Since each $k_\ell$ is a small constant and $t = O(\log_J M)$, the summary tokens contribute $O(\log M)$ and the raw tokens contribute $O(k_1 L)$. The attention cost is therefore $O(k_1 L + \log M)$.
 
\paragraph{Decoding: Total.}
Combining routing and attention, the per-step decoding cost is:
\begin{equation}
    O(\log n + k_1 L) = O(\log n),
\end{equation}
which is \emph{logarithmic} in the context length. Over a full sequence of $n$ decoding steps, the aggregate cost is $O(n \log n)$, which is \emph{log-linear}.

\section{Experiments on Passkey Retrieval}
\label{sec:passkey}
\begin{figure}[htbp]
     \centering
     \begin{subfigure}[b]{0.24\textwidth}
         \centering
         \includegraphics[width=\linewidth]{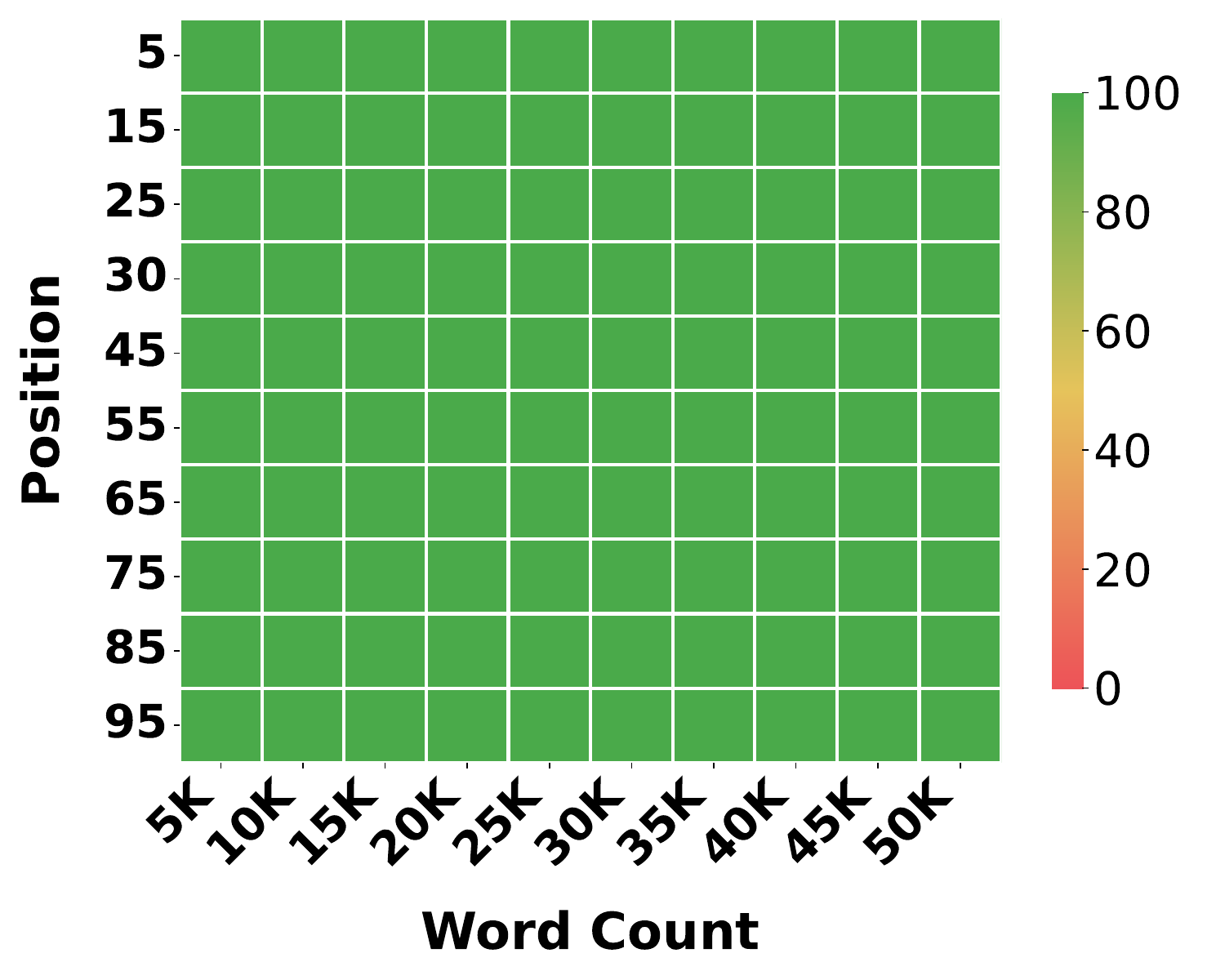}
         \caption{\toolName{} on Qwen2}
         \label{fig:task1}
     \end{subfigure}
     \begin{subfigure}[b]{0.24\textwidth}
         \centering
         \includegraphics[width=\linewidth]{task_heatmap_visualization_qwen.pdf}
         \caption{H-\toolName{} on Qwen2}
         \label{fig:task2}
     \end{subfigure}
     \begin{subfigure}[b]{0.24\textwidth}
         \centering
         \includegraphics[width=\linewidth]{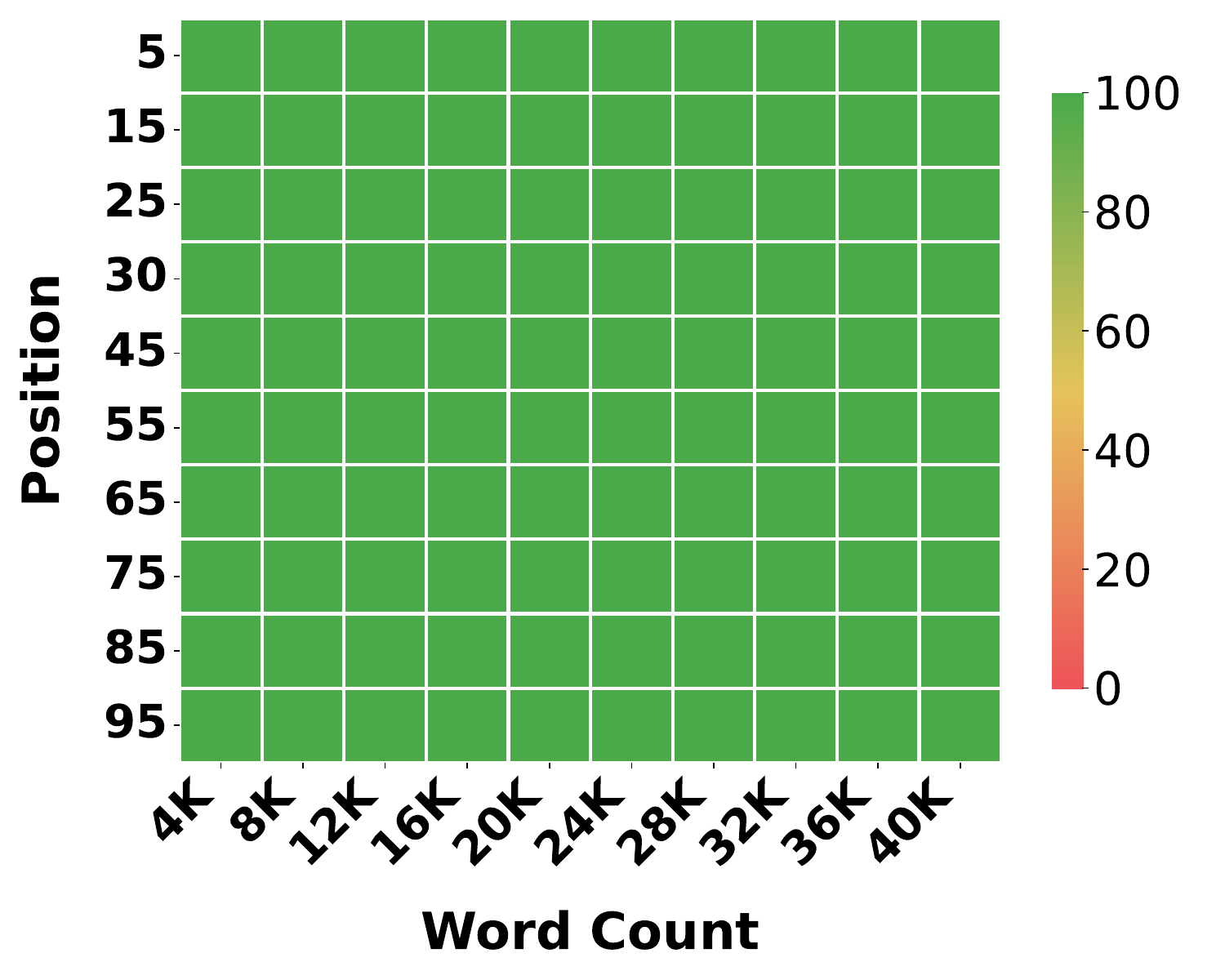}
         \caption{\toolName{} on Llama3.2}
         \label{fig:task3}
     \end{subfigure}
     \begin{subfigure}[b]{0.24\textwidth}
         \centering
         \includegraphics[width=\linewidth]{task_heatmap_visualization.pdf}
         \caption{H-\toolName{} on Llama3.2}
         \label{fig:task4}
     \end{subfigure}
     
     \caption{Passkey retrieval accuracy of \toolName{} and H-\toolName{} on Qwen2-7B-Instruct (left two panels) and Llama3.2-1B (right two panels). The vertical axis denotes the context length, while the horizontal axis represents the relative insertion position (\%) of the passkey.}
     \label{fig:passkey}
\end{figure}

To evaluate fine-grained retrieval under compression, we test on the passkey retrieval benchmark, where a short passkey is inserted at a random position within a long distractor context and the model must locate and reproduce it exactly. Figure~\ref{fig:passkey} reports retrieval accuracy as a function of context length (horizontal axis) and relative insertion position (vertical axis) for both after continue pretrained \toolName{} and H-\toolName{} on Qwen2-7B-Instruct (up to 50K words, $2.5\times$ the maximum sequence length seen during continued pretraining) and Llama-3.2-1B (up to 40K words, $10\times$ the maximum sequence length seen during continued pretraining).
 
Both \toolName{} and H-\toolName{} achieve \textbf{perfect 100\% accuracy} across all context lengths, all insertion positions, and both model families. The uniformly green heatmaps confirm that the selective unfolding mechanism reliably identifies and recovers the passkey regardless of where it is placed---beginning, middle, or end---and regardless of how long the surrounding context is. This holds even for the hierarchical variant, where the passkey must be located through multi-level coarse-to-fine routing. The result demonstrates that \toolName{}'s selective unfolding mechanism is robust and position-unbiased. It also demonstrates that \toolName{} generalizes robustly to sequence lengths far exceeding those seen during continued pretraining.

\end{document}